\documentclass[10pt,twocolumn,letterpaper]{article}

\usepackage{wacv}              

\usepackage{graphicx}
\usepackage{amsmath}
\usepackage{amssymb}
\usepackage{booktabs}

\usepackage{color}
\definecolor{Gray}{gray}{0.9}
\definecolor{Gray2}{gray}{0.95}
\usepackage{color,colortbl}
\definecolor{applegreen}{rgb}{0.0, 0.5, 0.0}
\definecolor{amethyst}{rgb}{0.6, 0.4, 0.8}
\usepackage{multirow}
\usepackage[accsupp]{axessibility}

\usepackage[pagebackref,breaklinks,colorlinks]{hyperref}

\usepackage[capitalize]{cleveref}
\crefname{section}{Sec.}{Secs.}
\Crefname{section}{Section}{Sections}
\Crefname{table}{Table}{Tables}
\crefname{table}{Tab.}{Tabs.}

\def\wacvPaperID{30} 
\def\confName{WACV}
\def\confYear{2025}

\begin{document}


\title{Local Masked Reconstruction for Efficient Self-Supervised Learning on High-resolution Images}

\author{
Jun Chen$^{1}$ $^*$,
Faizan Farooq Khan $^{1}$ $^*$, 
Ming Hu$^{2}$, 
Ammar Sherif$^{3}$, \\
Zongyuan Ge$^{2}$, 
Boyang Li$^{4}$,
Mohamed Elhoseiny$^{1}$ \\
\texttt{\{jun.chen, faizan.khan,mohamed.elhoseiny\}@kaust.edu.sa} \\
\texttt{\{ming.hu, zongyuan.ge\}@monash.edu} \\
\texttt{\{libo0001@gmail.com, asherif@nu.edu.eg\}} \\
$^{1}$King Abdullah University of Science and Technology\hspace{0.1cm} \\
$^{2}$Monash University \hspace{0.1cm}
$^{3}$Nile University  \hspace{0.1cm}
$^{4}$Nanyang Technological University  \hspace{0.1cm}
}
\maketitle
\def\thefootnote{*}\footnotemark\footnotetext{Equal Contribution}
\begin{abstract}
Self-supervised learning for computer vision has progressed tremendously and improved many downstream vision tasks, such as image classification, semantic segmentation, and object detection. Among these, generative self-supervised vision learning approaches, such as MAE and BEiT, show promising performance. However, their global reconstruction mechanism is computationally demanding, especially for high-resolution images. The computational cost increases extensively when scaled to a large-scale dataset. To address this issue, we propose local masked reconstruction (LoMaR), a simple yet effective approach that reconstructs image patches from small neighboring regions. The strategy can be easily integrated into any generative self-supervised learning techniques and improves the trade-off between efficiency and accuracy compared to reconstruction over the entire image. LoMaR is 2.5$\times$ faster than MAE and 5.0$\times$ faster than BEiT on 384$\times$384 ImageNet pretraining and surpasses them by 0.2\% and 0.8\% in accuracy, respectively. It is 2.1$\times$ faster than MAE on iNaturalist pretraining and gains 0.2\% in accuracy. On MS COCO, LoMaR outperforms MAE by 0.5 $\text{AP}^\text{box}$ on object detection and 0.5 $\text{AP}^\text{mask}$ on instance segmentation. It also outperforms MAE by 0.2\% on semantic segmentation. Our code and pretrained models are available at:  \href{https://github.com/junchen14/LoMaR}{https://github.com/junchen14/LoMaR}.
\end{abstract}

\section{Introduction}
\label{sec:intro}
Recently, self-supervised learning \cite{moco3,dino,beit,mae,igpt,bachman2019learning,wu2018unsupervised,oord2018representation,hjelm2018learning} has achieved enormous success in learning representations conducive to downstream applications, such as image classification and object detection. Among these, several generative methods, such as  Masked Autoencoder (MAE) \cite{mae} and Bidirectional Encoder Representation from Image Transformers (BEiT) \cite{beit}, which reconstruct the input image from a small portion of image patches, have demonstrated excellent performance. 

However, a significant bottleneck of MAE and BEiT is their high demand for compute, as they reconstruct masked image patches from global information and operate on a large number of image patches. For example, pretraining an MAE-Huge network on ImageNet under 224 $\times$ 224 resolution takes 34.5 hours on 128 TPU-v3 GPUs. BEiT\cite{beit} training is even slower due to the cost of the discrete variational autoencoder. 

High-resolution images further exacerbate this issue due to the $\mathcal{O}(n^2)$ time complexity of the Transformer model on $n$ image patches. For example, pretraining MAE on 384$\times$384 images consumes 4.7 times the compute time of 224$\times$224 counterpart. 
However, high-resolution images are essential in many tasks, such as object detection.
Thus, improving the efficiency of pretraining holds the promise to unleash additional performance gains under pretraining with a much larger dataset or higher-resolution images.

\begin{figure*}[t!]
\centering
\includegraphics[width=0.87\linewidth]
                  {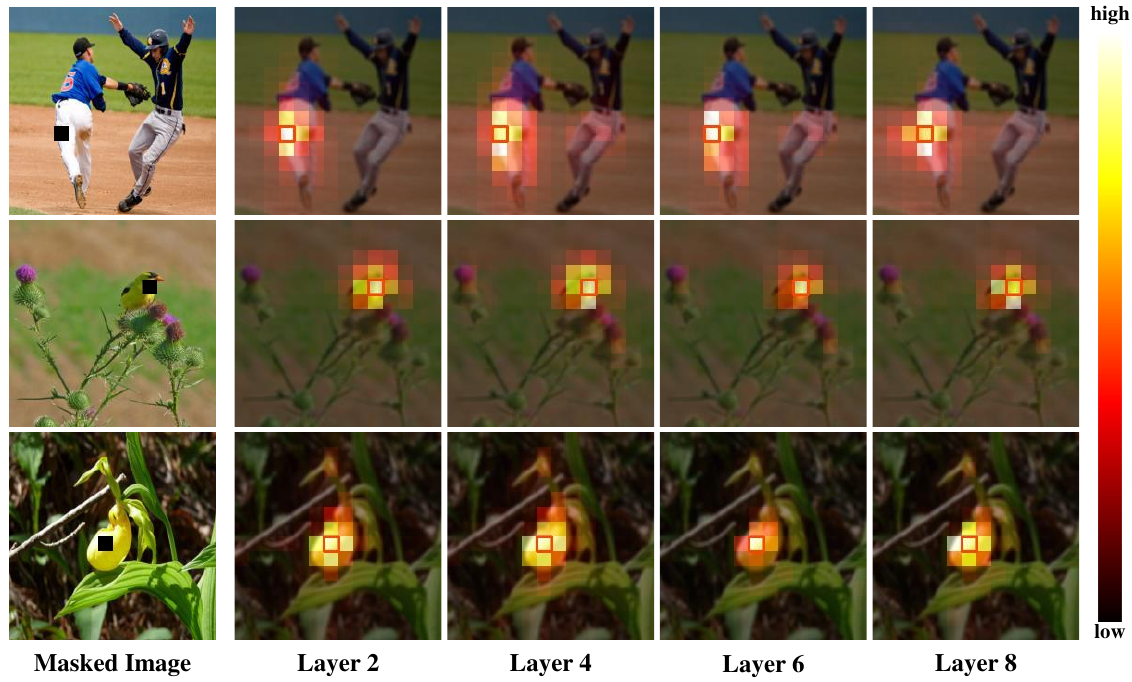}
\caption{We visualize the attention patterns employed by $\text{MAE}_{\text{Large}}$ \cite{mae} in the reconstruction of a random target patch, indicated by orange. Patches that are important for prediction are usually close to the target patch. We selected the images randomly from the ImageNet-1K \cite{imagenet} Val set. }
\label{fig:attention}
\end{figure*}

We observe that most reconstruction in MAE relies on local information only. 
In  Fig. \ref{fig:attention}, we visualize the attention weights (white indicating high attention) when reconstructing a target image patch. From a pretrained $\text{MAE}_{\text{Large}}$ model, we extract the attention weights from the decoder layers 2, 4, 6, and 8.  The model mainly attends to patches close to the target patch, which motivates us to limit the range of attention used in the reconstruction. 

Hence, we propose a new model, dubbed \textbf{Lo}cal \textbf{Ma}sked \textbf{R}econstruction or LoMaR. The model restricts the attention region to a small window, such as 7$\times$7 image patches, which is sufficient for reconstruction. 
Similar approaches \cite{sparsetransformer,attentionspan,bigbird} have been seen in NLP problems that need to process long sequences. The small windows have also been explored in vision domains for higher training and inference speed \cite{swintransformer,focalTransformer}. However, unlike prior work in vision transformers, which create shifting windows with fixed coordinates for each image, we sample several windows with random locations, which can better capture the objects in different spatial areas.

\begin{figure*}[t!]
\centering
 \includegraphics[width=1.0\linewidth]
         {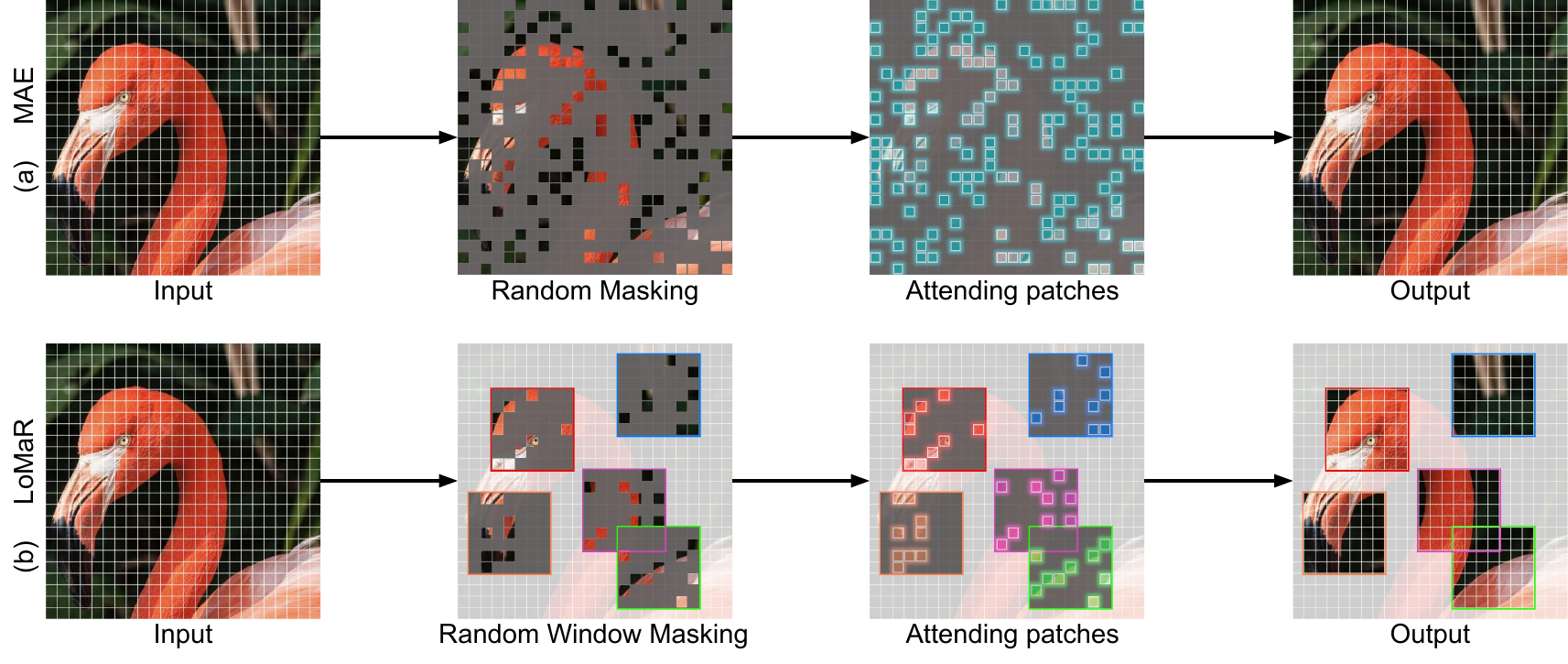}
\caption{\textbf{Contrasting the masking and reconstruction strategy between MAE and LoMaR.} During the pretraining, MAE randomly masks 75\% patches as masking and reconstructs them by attending to the remaining visible patches. For LoMaR, it randomly samples several small regions and masks a random subset of patches from each region, e.g. 80\%. The masked patches will only attend to the visible patches inside each region for reconstruction. In contrast to MAE, LoMaR usually samples less visible patches per image.}
\label{architecture}
\end{figure*}

In Figure \ref{architecture}, we compare LoMaR and MAE and note two significant differences: a) We sample a region with k$\times$k patches to perform masked reconstruction instead of from the total number of patches.
Instead of reconstructing the masked patches from the 25\% visible patches globally located in the image, we find that it is sufficient to recover the missing information with only some local visual clues.
b) We replace the heavy-weight decoder in MAE with a lightweight MLP head. We feed all image patches directly into the encoder, including masked and visible patches. In comparison, only the visible patches are fed to the encoder in MAE. Experiments show that these architectural changes bring more performance gain to the local masked reconstruction in small regions.

After conducting extensive experiments, we found that 
\begin{itemize}
    \item LoMaR is more efficient than other baselines in pretraining on high-resolution images since its computation cost is invariant to the different image resolutions. However, other approaches have a quadratic computational cost to the image resolution increase, which leads to much more expensive pretraining. For example, for pretraining on 448$\times$448 images, LoMaR is 3.1$\times$ faster than MAE and 5.3 $\times$ faster than BEiT while achieving higher classification performance.
    \item LoMaR also has a strong generalization ability on object detection and semantic segmentation tasks. It outperforms MAE by 0.5 $\text{AP}^\text{box}$ under ViTDet \cite{li2022exploring} framework for object detection. Also, it outperforms MAE by 0.2 points under UperNet \cite{upernet} for semantic segmentation.
    \item LoMaR is efficient and can be easily integrated into any other generative self-supervised learning approach. Equipping our local masked reconstruction learning mechanism into BEiT can improve its ImageNet-1K classification performance from 83.2 to 83.4 Top-1 accuracy, costing only 35.8\% of its original pretraining time. 
\end{itemize}

\section{Related Work}
\label{sec:related}

\noindent \textbf{Self-supervised Learning For Images.} The past few years have witnessed a boom in self-supervised learning. Existing techniques can be broadly categorized into discriminative \cite{becker1992self} and generative \cite{pathak2017learning,gidaris2018unsupervised}. The prominent discriminative approach, instance discrimination, distinguishes different views of the same data instance from other instances \cite{wu2018unsupervised,oord2018representation,he2020momentum,chen2020simple,grill2020bootstrap,chen2021exploring}. 
The most representative works include BYOL \cite{grill2020bootstrap} , MOCO \cite{chen2020improved,moco3,he2020momentum} and SimCLR \cite{chen2020simple,chen2020big}. VICRegL\cite{bardes2022vicregl} performs the contrasting learning in local and global features. Other approaches such as SwAV \cite{caron2020unsupervised} and DINO \cite{dino} heavily rely on multi-crop strategy, for instance, discrimination. The generative approach includes autoregressive prediction \cite{han2019video,igpt} and autoencoders, which we discuss next. 

\noindent \textbf{Autoencoders for Representation Learning.} 
An autoencoder, which aims to learn a representation from which the original input can be reconstructed, has been a popular choice for representation learning since the dawn of deep learning \cite{Hinton2006:DeepAutoEncoder,Bengio2006:Layerwise-Training,bengio2013representation}. The autoencoding problem is inherently ill-posed due to the existence of a trivial solution: a network entirely composed of identity mappings. Hence, it is necessary to apply some form of regularization, such as sparsity \cite{Ranzato2006:SparseAutoEncoder}, input corruption \cite{Vincent2008:DenoisingAutoencoders}, probability priors \cite{kingma2013:VariationalBayes,Rezende2014:VAE}, or adversarial discriminators \cite{makhzani2015adversarial}.  

In particular, denoising autoencoder \cite{Vincent2008:DenoisingAutoencoders}, which attempts to recover the original input from a corrupted version, has received significant research attention. Variations include solving jigsaw puzzles \cite{noroozi2016jigsaw}, color restoration \cite{zhang2016colorful,larsson2016learning},  spatial relation recovery \cite{doersch2015:spatial-relation}, inpainting \cite{pathak2016:impainting}, and so on. Recently, BEiT \cite{beit} proposes to encode image patches as a dictionary using dVAE \cite{ramesh2021zero} and predict the encoding of missing patches. PeCo \cite{peco} further improves BEiT by enforcing perceptual similarity from dVAE. MAE \cite{mae} reconstructs directly the missing pixels. CiM \cite{cim} replaces image patches with plausible alternatives and learns to recover the original and predict which patches are replaced. Data2vec \cite{data2vec} performs self-supervised learning across multi-modalities. MultiMAE~\cite{bachmann2022multimae} shows that multi-modality pretraining can be more training-efficient than single-modality. Unlike prior methods, LoMaR achieves linear complexity by restricting self-attention to localized patches, rather than relying on global reconstructions with quadratic complexity.

\section{Approach}
\label{sec:approach}
LoMaR relies on a stack of Transformer \cite{transformer} blocks to pre-train on a large number of unlabeled images by recovering the missing patches from corrupted images, similar to MAE \cite{mae}. Still, LoMaR differentiates from MAE in several key places.  
In this section, we first revisit the MAE model and then describe the differences between LoMaR and MAE.

\subsection{Background: Masked Autoencoder} The Masked Autoencoder (MAE) model \cite{mae}, employs an asymmetric encoder-decoder architecture. The encoder takes in a subset of patches from an image and outputs latent representations for the patches. From those, the decoder reconstructs the missing patches. For an input image with resolution $h$ $\times$ $w$, MAE first divides it into a sequence of non-overlapping patches. Then, MAE randomly masks a large proportion (e.g., 75\%) of image patches; see the upper side of Fig. \ref{architecture}. The positional encodings are added to each patch to indicate their spatial location. MAE first encodes the remaining patches into the latent representation space. Then, it feeds the latent representations with placeholders for the masked patches into the decoder, which carries out the reconstruction. For each reconstructed image, MAE uses the mean squared error (MSE) with the original image in the pixel space as the loss function.

\begin{table*}[h!]
\begin{center}
\begin{tabular}{lccccccccc}
\toprule
& & \multicolumn{3}{c}{ImageNet 1K} &\multicolumn{3}{c}{Inaturalist} \\
Method & Resolution &  Time (h) $\downarrow$ & Top-1 Acc& Speed-up & Time (h) $\downarrow$ & Top-1 Acc & Speed-up\\
\midrule
BEiT \cite{beit} & 384& $\sim$408 & 83.7 &1.0$\times$  & $\sim$93 & 81.1 & 1.0$\times$ \\
MAE & 384& $\sim$203 & 84.3 & 2$\times$ & $\sim$46  &81.2 & 2.0$\times$\\
\rowcolor{Gray2}
LoMaR (ours) & 384& \textbf{$\sim$81} & \textbf{84.5}  & \textbf{5.0$\times$} & \textbf{$\sim$22}& \textbf{81.4} & \textbf{4.2$\times$} \\
\midrule
BEiT \cite{beit} & 448& $\sim$595 & 84.1 & 1.0$\times$& $\sim$121 & 82.3 & 1.0$\times$\\
MAE & 448& $\sim$345 & 84.5 & 1.7$\times$ & $\sim$70 & \textbf{82.4} & 1.7$\times$ \\
\rowcolor{Gray2}
LoMaR (ours) & 448& \textbf{$\sim$113} & \textbf{84.7} & \textbf{5.3$\times$} & \textbf{$\sim$32} & 82.3 & \textbf{3.8$\times$}\\
\bottomrule
\end{tabular}
\end{center}
\caption{High-resolution image pretraining and classification results on ImageNet-1K dataset \cite{imagenet} and Inaturalist \cite{van2018inaturalist}. The pretraining times are all computed on 4 NVIDIA 80GB A100 GPUs. We take BEiT~\cite{beit} as the comparison baseline when computing the speed-up for MAE~\cite{mae} and LoMaR. Our LoMaR can always achieve comparable or higher performance with at least a 3.8$\times$ speed-up than BEiT and a 2.2$\times$ speed-up than MAE on both datasets.}
\label{high_resolution}
\end{table*}

\subsection{Local Masked Reconstruction (LoMaR)}

We describe LoMaR by contrasting it with MAE from the following perspectives.

\noindent \textbf{Local vs. Global Masked Reconstruction.} MAE reconstructs each missing patch with patches sampled from the entire image. 
However, as indicated by Fig \ref{fig:attention}, usually only the patches in the proximity of the target patch contribute significantly to the reconstruction, suggesting that local information is sufficient for reconstruction. Therefore, we perform the random window masking and reconstruction on patches within a small region, shown in the bottom side of Fig. \ref{architecture}. Specifically, we perform the random window masking by sampling several small regions from the image and restricting the masked patches to only attend to their local surrounding visible patches, as we highlighted. Our experiments find that a region size of 7$\times$7 patches leads to the best trade-off between accuracy and efficiency. 
On the other hand, similar to convolutional networks \cite{he2016deep,VGG}, LoMaR has the translation invariance property due to the usage of small windows sampled in random spatial locations each iteration.

From the complexity perspective, local masking and reconstruction are more computationally efficient than global masking and reconstruction of MAE because there are fewer tokens for operation. Suppose each image can be divided into $h \times w$ patches. The time complexity for computing the self-attention is $\mathcal{O}(h^2w^2)$. The complexity is quadratic to the number of patches and hard to scale up with large $hw$. However, for our local masked reconstruction, we sample $n$ windows where each contains $m \times m$ patches; Its computational complexity is $\mathcal{O}(nm^4)$, which has linear time complexity if we fix $m\times m$ as a constant window size. It can reduce the computational cost significantly if $nm^4 \ll h^2w^2$. 
For example, for a 448x448 image, the cost of self-attention calculation is reduced from $448^2$ x $448^2$ in the case of MAE to $4$x$7^2$x$7^2$ in the case of LoMaR when we sample 4 views of 7$\times$7 patches.

\noindent \textbf{Architecture.} Instead of the asymmetric encoder-decoder of MAE, LoMaR only applies a simple Transformer encoder architecture.
We input all the visible and masked patches under a sampled region into the encoder and reconstruct the masked patches through a simple MLP layer. Although feeding the masked patches into the encoder can be deemed a less efficient operation than MAE, which only inputs masked patches into the decoder, we find that inputting the masks in the early stages can enhance the visual representation and make it more robust to mask reconstruction from the smaller regions. It might be because the encoder can convert the masked patches back to their original RGB representation after multiple encoder layers interact with the other visible patches. Those recovered masks in the hidden layers can implicitly contribute to the image representation. Therefore, LoMaR preserves the mask patches as the encoder input.

\noindent \textbf{Implementation.} 
Given an image, we first divide it into several non-overlapping patches. Each patch is linearly projected into an embedding. We randomly sample several square-shaped regions of $K \times K$ patches at different spatial locations. We then zeroed out a fixed percentage of patches within each region. After that, we feed all the patches, including visible and masked ones, from each region to the encoder in raster order. 
We also apply the relative positional encoding \cite{wu2021rethinking} into our model, enabling the translation-invariant property for the local masked reconstruction. We convert the latent representations from the encoder output to their original feature dimension with a simple MLP head and then compute the mean squared error with the normalized ground-truth image.

\section{Experiments}
\label{sec:exp}
We examine the performance of LoMaR by pretraining and finetuning on ImageNet-1K \cite{imagenet} dataset with the following procedure. First, we perform the self-supervised pretraining on the ImageNet-1K training dataset without label information. Then, we finetune the pre-trained model on ImageNet-1K with supervision from the labels. During finetuning, we feed all the image patches to the model and take the average of their features as the final representation for classification. We follow the same experimental settings as MAE \cite{mae}; detailed hyperparameters can be found in the supplementary material.

\subsection{Experiments on High-resolution Images} We evaluate our model, MAE \cite{mae} and BEiT \cite{beit}  on ImageNet \cite{imagenet} and Inaturalist \cite{van2018inaturalist} datasets. We pre-train and finetune on high-resolution images such as 384 $\times$ 384 and 448 $\times$ 448 images. We follow MAE's default settings during pretraining; Sample 75\% patches as masks. For LoMaR, we set the number of views to 6 and 9 for resolutions of 384 and 448 on the ImageNet dataset and sample 8 and 12 views for resolutions of 384 and 448 on the iNaturalist dataset. We pretrain all the models with 300 epochs and finetune them under the same image resolution. 

We summarize the results in Table \ref{high_resolution}. The results demonstrate that LoMaR outperforms other models with substantially less pretraining time, which scales linearly with the window numbers. In contrast, the pretraining time of MAE and BEiT scales quadratically as the resolution increases. As a result, LoMaR is 2.5$\times$ faster than MAE (\textcolor{applegreen}{accuracy +0.2\%}) and 5.0$\times$ faster than BEiT (\textcolor{applegreen}{accuracy +0.8\%}) on 384$\times$384 images, and for the resolution of 448$\times$448, it is 3.1$\times$ faster than MAE (\textcolor{applegreen}{accuracy +0.2\%}) and 5.3$\times$ faster than BEiT (\textcolor{applegreen}{accuracy +0.6\%}). On Inaturalist, LoMaR is 2.1$\times$ faster than MAE model (\textcolor{applegreen}{accuracy +0.2\%}) and 4.2$\times$ faster than BEiT (\textcolor{applegreen}{accuracy +0.3\%}), and it can produce comparable performance with the other baselines but is 3.8$\times$ faster than BEiT and 2.2$\times$ faster than MAE.

\begin{table}[t!]
\small
\setlength{\tabcolsep}{3pt}
\begin{center}
\begin{tabular}{lccccc}
\toprule
Methods & Epochs & Res & Time (h) $\downarrow$& Top-1 Acc & Speed-up\\
\midrule
No Pretraining  & -  & 224 & - &82.3 &- \\
DINO \cite{dino}   & 300  & 224 &- & 82.8 &- \\
MoCov3 \cite{moco3} & 600 & 224 &-&   83.2&- \\
MSN~\cite{assran2022masked} & 600 & 224 & - &  83.4 & - \\
CAE~\cite{ContextAutoencoder2022} & 300 & 224 & - & 83.6 & - \\
CAE~\cite{ContextAutoencoder2022} & 800 & 224 & - & 83.8 & - \\
CAE~\cite{ContextAutoencoder2022} & 1600 & 224 & - & 83.9 & - \\
MMAE~\cite{mmae} & 1600 & 224 & - & 83.3 & - \\

SemMAE~\cite{li2022semmae} & 800 & 224 & - & 83.3 & - \\
\midrule    
BEiT \cite{beit}  \cite{ramesh2021zero}  & 300 & 224& $\sim$107 &82.9  & 1.0$\times$ \\

MAE* \cite{mae}   &400 & 224 &$\sim$58&83.1 & 1.8$\times$\\

\rowcolor{Gray2}
LoMaR &300 & 224 & $\sim$49& 83.3 & 2.2$\times$\\
\midrule
$\text{LoMaR}_{8\times 8}$ & 400 & 224 &$\sim$66& 84.3 & 1.6$\times$\\
\bottomrule
\end{tabular}
\end{center}
\caption{Image classification results on the ImageNet-1K (IN1K) dataset \cite{imagenet}. All baselines excluding $\text{LoMaR}_{8\times 8}$ adopt the ViT B/16 model \cite{vit} and are pretrained on 224$\times$224 images. $\text{LoMaR}_{8\times 8}$ applies ViT B/8 as the backbone. * denotes our reproduced results based on the officially released code and pretrained models. The pretraining times are all computed on 4 NVIDIA 80GB A100 GPUs.}
\label{imagenet_results}
\end{table}

\subsection{Experiments on Low-resolution images }
\begin{figure}
    \centering
    \includegraphics[width=0.48\textwidth]{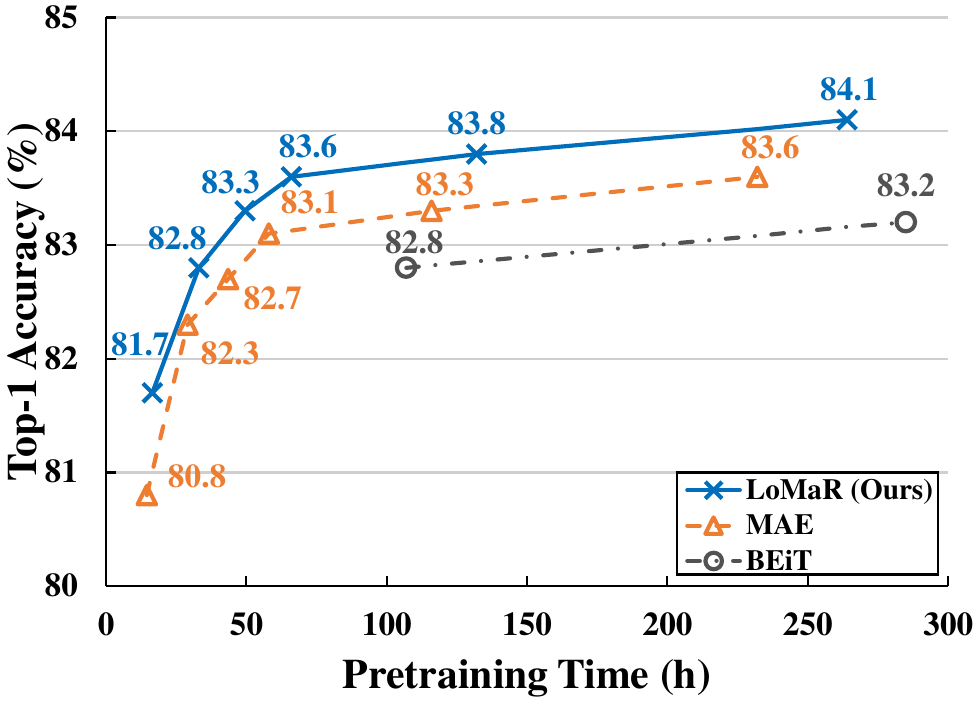}
\caption{\textbf{Computational efficiency evaluation}: We compute their ImageNet-1K top-1 accuracy per pretraining time for low-resolution images 224$\times$224.
  }
  \label{efficiency_evaluation}
\end{figure}

Table \ref{imagenet_results} summarizes the results of different self-supervised learning approaches. All models are pretrained in self-supervised fashion on ImageNet-1K \cite{imagenet} under the 224$\times$224 resolution and finetuned on labeled ImageNet-1K. LoMaR reaches the best result of MAE, 83.6\%, after only 400 epochs of pretraining. When pretrained for 1,600 epochs, its performance further improves to 84.1\%; this can also be seen in Fig.~\ref{efficiency_evaluation} after $\approx$250 hours of pre-training. When finetuned under the 384$\times$384 resolution, LoMaR reaches an accuracy of 85.4\%, 0.6\% higher than the best baseline. Overall, LoMaR outperforms strong baselines with less pretraining time. 

\textbf{Efficiency analysis.} We train LoMaR, MAE \cite{mae} and BEiT \cite{beit} baselines with different pretraining epochs [100, 200, 300, 400, 800, 1,600] on 224$\times$224 images. We compare their pretraining time v.s top-1 accuracy in Fig. \ref{efficiency_evaluation}.  We carefully tuned all models to achieve the best load balancing between the GPU and the CPU and the maximal image throughput during training. We do this by adjusting ghost batch size \cite{Hoffer2017} while keeping the total batch size constant for all models. Compared to baselines, we observe that LoMaR consistently achieves the same or higher accuracy in less pretraining time. Specifically, pretraining MAE achieves 83.6\% accuracy but takes about 232 hours. LoMaR reaches the same accuracy within $\sim$66 hours of pretraining, 3.5$\times$ faster. BEiT requires $285$ pretraining hours to get 83.2\% accuracy. In contrast, LoMaR obtains a similar result within $\sim$49 hours, translating to 5.8$\times$ time savings.

CAE~\cite{ContextAutoencoder2022}, MMAE~\cite{mmae}, and SEMMAE~\cite{li2022semmae} build on top of MAE~\cite{mae} architecture adding additional components. At best, the time complexity of these approaches is similar to that of MAE, so for high-resolution pre-training, LoMaR is much more efficient. MSN~\cite{assran2022masked} differs from MAE as it matches the representation of an image with randomly masked patches to the representation of the original unmasked image. We calculate the time per epoch for MSN at a high resolution of 384 and 448 and compare it with LoMaR; for 384, MSN takes 0.64 hours/epoch, while for LoMar, it takes 0.27 hours/epoch. For 448, MSN takes 0.88 hours/epoch, while LoMar takes 0.38 hours/epoch. This comparison also demonstrates the computational efficiency of our LoMaR model.

\begin{table}[t!]
\begin{center}

\begin{tabular}{llllll}
\toprule
\multirow{2}{*}{Backbone} & \multicolumn{2}{c}{ViTDet*\cite{li2022exploring}} & \multicolumn{2}{c}{ViTAE \cite{zhang2022vitaev2}} \\
 & $\text{AP}^{\text{box}}$ & $\text{AP}^{\text{mask}}$ & $\text{AP}^{\text{box}}$ & $\text{AP}^{\text{mask}}$ \\
\midrule
MAE & $\text{51.1}^*$  & $\text{45.4}^*$ & 51.6 & 45.8 \\
\rowcolor{Gray2}
LoMaR & 51.4  & 45.7 & 51.8 & 46.0 \\
\rowcolor{Gray2}
$\text{LoMaR}_{\text{384}}$ & \textbf{51.6}  &  \textbf{45.9} & \textbf{52.0} & \textbf{46.2}\\
\bottomrule
\end{tabular}
\end{center}
\caption{\textbf{Object detection and instance segmentation results} on MS COCO under two ViT frameworks. * denotes reproduced results with the code from \cite{zhang2022vitaev2}. $\text{LoMaR}_{\text{384}}$ denotes the model pretrained on 384$\times$384 images. Other models are pretrained on 224$\times$224 images.}
\label{object_detection}
\end{table}

\textbf{Pretraining on small patches.} We also evaluate our model on smaller patches with 8$\times$8 pixels instead of the usual 16$\times$16 pixels. We employ ViT B/8 \cite{vit} as a backbone in Table \ref{imagenet_results}. We pre-train LoMaR with 7$\times$7 windows (4 views per image) on 224$\times$224 images for 400 epochs. It is worth noting that this incurs the same computation time (about 66 hours) as 16$\times$16 patches. The model accuracy after finetuning reaches to 84.3\% top-1 accuracy. However, similar experiments are costly for MAE and BEiT, as smaller patches substantially increase the number of patches for operation and lead to the high cost of self-attention. In our experiments, the pretraining of MAE with their official code under smaller patches crashes due to numerical issues.

\subsection{Object Detection and Instance Segmentation.} We finetune our model end-to-end on MS COCO \cite{mscoco} for the object detection and instance segmentation tasks. We replace the ViT backbone with our pretrained LoMaR model in the ViTDet \cite{li2022exploring} and ViTAE \cite{zhang2022vitaev2} frameworks. We report object detection results in $\text{AP}^{\text{box}}$ and instance segmentation results in $\text{AP}^{\text{mask}}$. 

We provide the results in Table \ref{object_detection}. It shows the consistent improvement of LoMaR on the COCO object detection benchmark. Under ViTDet, LoMaR surpasses  MAE by 0.3 $\text{AP}^{\text{box}}$ and 0.3 $\text{AP}^{\text{mask}}$. When applying the LoMaR pretrained for 1,600 epochs under the 384$\times$384 resolution, it further improves to 51.6 $\text{AP}^{\text{box}}$ and 45.9 $\text{AP}^{\text{mask}}$. In the ViTAE framework, LoMaR improves over MAE by 0.4 $\text{AP}^{\text{box}}$ and 0.4 $\text{AP}^{\text{mask}}$, respectively.

\begin{table}[t!]
\begin{center}
\begin{tabular}{lcc}
\toprule
Models & Pre-train Data & ADE20K\\
\midrule
supervised &IN1K w/ labels & 47.4\\
BEiT~\cite{beit} &IN1K+DALLE & 47.1\\
MAE~\cite{mae}& IN1K & 48.1\\
LoMaR &IN1K & 47.8\\
\rowcolor{Gray2}
$\text{LoMaR}_{\text{384}}$ &IN1K & \textbf{48.3}\\
\bottomrule
\end{tabular}
\end{center}
\caption{\textbf{Semantic segmentation on ADE20K \cite{ade20k} (mIoU)}. All the baselines are computed under UperNet \cite{upernet} framework. $\text{LoMaR}_{\text{384}}$ denotes the results under pretraining images with the resolution of 384$\times$384. Other baselines are pretrained on 224$\times$224 images.}
\label{sementic_segmentation}
\end{table}

\subsection{Semantic Segmentation} We evaluate our model on the semantic segmentation benchmark, ADE20K \cite{ade20k}, and compare with the baselines in Table \ref{sementic_segmentation}. We train the UperNet \cite{upernet} model with our pretrained LoMaR as initialization. When applying the LoMaR pretrained on images with 384$\times$384 resolution, it consistently outperforms MAE by 0.2 points. This shows the consistent improvement of our LoMaR over the MAE and BEiT baselines. Additionally, it demonstrates the usefulness of high-resolution image pretraining.

\begin{table}[]
    \centering
    \begin{tabular}{cccc} 
    \hline
    Method & Time(h)$\downarrow$ & Top-1 Acc & Speed-up \\
    \hline
    BEiT &  $\sim$285 & 83.2 & 1$\times$\\
    \rowcolor{Gray2}
    BEiT+ window masking   & $\sim$102 & \textbf{83.4} & 2.8$\times$ \\\bottomrule
    \end{tabular}
\caption{The results of applying our method on the BEiT approach.}
\label{beit_adaption}
\end{table}

\subsection{Integration to BEiT}
Our core idea, local masked reconstruction, can be easily integrated into other generative self-supervised learning methods. To examine its effectiveness in a different paradigm, we integrate it to BEiT \cite{beit}. Specifically, we randomly sample four 7$\times$7 windows, feed them into the BEiT model, and pretrain for 300 epochs while retaining all other experimental settings as the original BEiT. Results in Table \ref{beit_adaption} show that this strategy improves the accuracy from 82.8\% to 83.4\%, which is higher than the original BEiT, and speeds up the training by 2.8$\times$.

\begin{figure}
    \centering
    \includegraphics[width=0.48\textwidth]{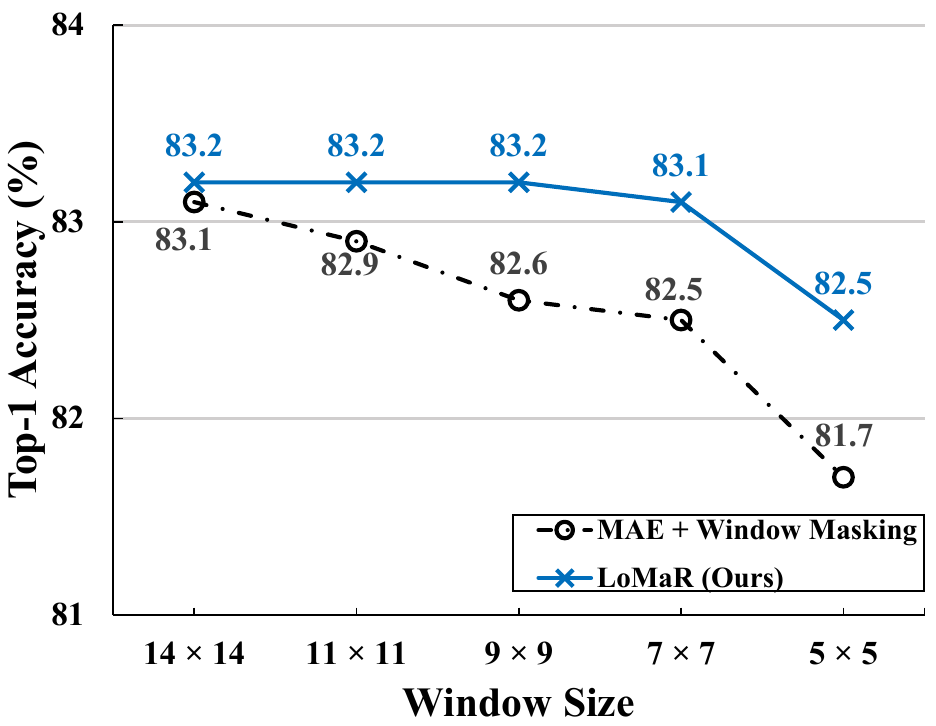}
\caption{Comparison between LoMaR simple encoder and MAE asymmetric encoder-decoder architectures on our random window masking strategy. The window sizes vary from 14$\times$14 to 5$\times$5.}
\label{ablation}
\end{figure}

\subsection{Ablative Experiments}
We conducted many ablation experiments to explore properties such as the window size, masking ratio, and architecture design and share our findings in this section. We performed all the ablation experiments under 4 NVIDIA 80GB A100 GPUs with the same setting for fair comparisons, and all the experiments were obtained by pretraining on 224$\times$224 images.

\noindent \textbf{Architecture.} Fig. \ref{ablation} compares different architectures, including a simple encoder (with both visible and masked patches as input) and MAE, an asymmetric encoder-decoder architecture with a local window. Initially, we sample 75\% patches as the masks following the guidance of MAE. 
By default, we use absolute positional encoding (APE) for both architectures.
We ablate these two architectures with different masked reconstruction windows, showing that a simple encoder can consistently outperform the asymmetric encoder-decoder. Moreover, the performance gap is further magnified when we decrease the window size from 14 to 7. This suggests a simple encoder is more robust to smaller window sizes than MAE-like architecture.

\begin{table}[]
    \centering
  \begin{tabular}{cccccc} 
    \hline
    Window Size & 5$\times$5 & 7$\times$7 & 9$\times$9 & 11$\times$11 & 14$\times$14\\
    \hline
    Views & 8 & 4 & 3 & 2 & 1\\
    \hline
  \end{tabular}
\caption{\textbf{Window size utilized :} The number of views per image, as utilized by LoMaR for different window sizes.}
\label{window_size}
\end{table}

\noindent \textbf{Efficiency vs. window size.} We test LoMaR with multiple different window sizes such as 5$\times$5, 7$\times$7, 9$\times$9, 11$\times$11 and 14$\times$14. One caveat is that the smaller window covers much fewer visible patches than the larger ones, which creates unfair comparisons. To encourage fairness, we assign different numbers of views for each window size, as we demonstrated in Table~\ref{window_size}. Thereby, all conditions have a similar number of visible patches in training. 

From the results in Fig.~\ref{ablation}, we can observe that while there is a performance drop when decreasing the window size from 14 $\rightarrow$ 5, the performance does not change much for other sizes, only 83.2 $\rightarrow$ 83.1, when decreasing the region size from 14$\times$ to 7. However, the restricted attention region decreases the total pretraining time from 120 to 66 hours. This means that pretraining on 7$\times$7 window size can roughly 2$\times$ speed up the pretraining process with minimal performance change. Therefore, window size 7$\times$7 can be deemed an optimal trade-off for local masked reconstruction.

\noindent \textbf{RPE vs. APE.} Relative positional encoding (RPE) has been widely used in previous works, including  BEiT~\cite{beit}. 
We also employ the RPE~\cite{wu2021rethinking} in LoMaR. We observe that it can bring 0.4 top-1 accuracy gain from 83.1 to 83.5. Therefore, we set RPE as the default setting for LoMaR in future experiments. 

\begin{figure*}[t!]
\centering
\includegraphics[width=0.94\linewidth]
                  {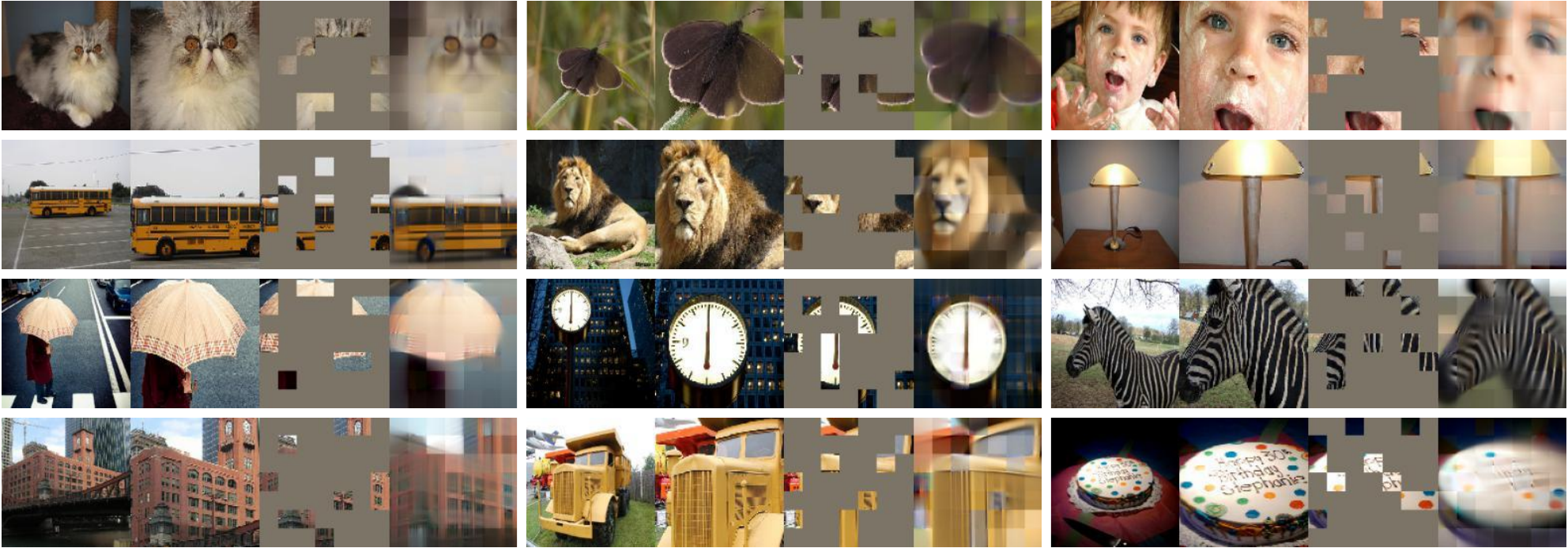}
\caption{Example results on ImageNet (upper two rows) and COCO (lower two rows) validation images. We mask 80\% patches out and reconstruct them with our pretrained model. For each image reconstruction figure, we split them into 4 parts: 1) the left-most is the original image. 2) the second-left is the sampled window (7$\times$7 patches). 3) The second-right is the masked image. 4) The right-most is our reconstructed image.}
\label{reconstruction}
\end{figure*}

\begin{figure}
    \centering
  \includegraphics[width=0.48\textwidth]{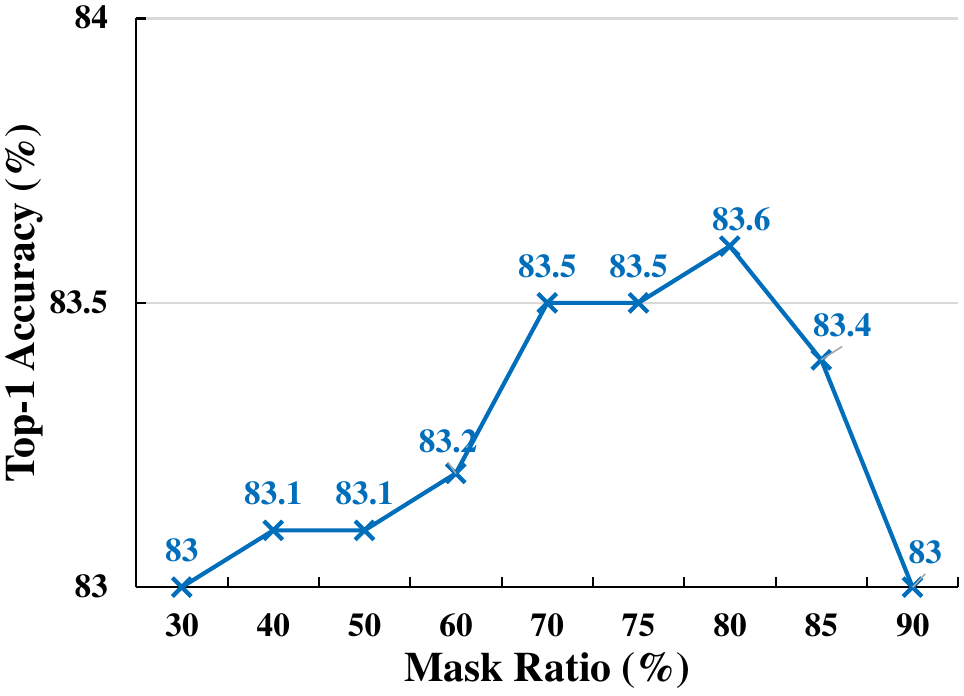}
\caption{\textbf{Mask ratio ablations}: It compares the LoMaR under different mask ratios from 30\% to 90\%}
\label{mask_ratio}
\end{figure}

\noindent \textbf{Mask ratio.} We also explore the best mask ratio under the local masked reconstruction scenario (see Fig. \ref{mask_ratio}). We train the previous best setting of our LoMaR on different mask ratios, ranging from 30\% to 90\%. The results show that too low (30\%) or too high (90\%) mask ratios are not optimal since they over-simplify or complicate the training task. We found that the 80\% mask ratio can result in the best performance, differentiating from the 60\% mask ratio observed in MAE for best finetuning performance. With this motivation, we employ the 80\% mask ratio in the rest of our experiments.

\subsection{Visualization of Reconstructed Images}

We qualitatively show the reconstruction performance of our pre-trained model in Fig. \ref{reconstruction}. We randomly sample several images from ImageNet-1K \cite{imagenet} and MS COCO\cite{mscoco}. After that, we sample a region containing 7$\times$7 patches in every image and zero out 80\% patches in the window for reconstruction. It can be seen that LoMaR is capable of generating plausible images, which also confirms our initial conjecture that the missing patches can be recovered from the local surrounding patches alone.



\section{Discussion and Limitation}
\label{sec:dis}

Self-supervised learning (SSL) can benefit from training with a massive amount of unlabeled data, which has brought many promising results \cite{bert,gpt1,gpt2,gpt3,mae,beit,moco3}. However, their high computational demands remain a significant concern under large-scale pretraining. In our study, we observe that the local masked reconstruction (LoMaR) for generative SSL is more efficient than the global version used by the influential works of MAE \cite{mae} and BEiT \cite{beit}. LoMaR demonstrates good generalization in image classification, instance segmentation, and object detection; it can be easily incorporated into both MAE and BEiT. LoMaR holds the promise to scale up SSL to even bigger datasets and higher resolution \cite{clip,JFT} datasets. LoMaR can also be extended to video analysis, where the computation problem is even more severe.

Another advantage of LoMaR is its efficiency gains when the number of image patches increases in high-resolution images such as 384$\times$384 and 448$\times$448 or even larger. The primary reason is that LoMaR restricts self-attention within a small region, and its computational complexity grows linearly with the number of sampled regions per image.
This characteristic enables efficient pretraining under high image resolution, which would be prohibitively expensive for other SSL methods. It can benefit many vision tasks such as object detection or instance segmentation, which require dense prediction at the pixel level.

Despite the high pretraining efficiency gain of LoMaR over other baselines for high-resolution images, one limitation is that LoMaR underperforms in linear probing (see results in Supplementary), which is mainly due to two reasons: 1)
There is a discrepancy between training and inference. During pretraining, we feed only a small region of patches and masked tokens to the network. The input contains all image patches without masked tokens during linear probing, resulting in a shift of input distribution and damaging linear probing performance. 2) LoMaR applies a much shallower decoder than MAE. A deep decoder improves linear probing performance because the last few layers in an autoencoder are specialized for reconstruction and not very helpful for recognition; MAE removes these layers during linear probing. However, as shown in Table \ref{imagenet_results}, fine-tuning the entire model can easily mitigate this limitation.
We hope the local masked reconstruction idea, pioneered by LoMaR, can lead to further research on efficient self-supervised learning.

{\small
\bibliographystyle{ieee_fullname}
\bibliography{egbib}
}

\appendix
\clearpage
\setcounter{page}{1}

\section{Experimental Details}

\subsection{ImageNet Experiments}
We launch our experiments by following the MAE \cite{mae} settings. The major implementation difference is the model implementation and positional encoding. We apply a 12-layer ViT \cite{vit} as a backbone. On top of the ViT, LoMaR adds an MLP layer for the linear projection. We also employ the relative positional encoding \cite{wu2021rethinking} in our LoMaR to model the relative positional relation among the patches within the sampled local window.

\textbf{Pretraining.} We apply the batch size 4096 by default during the pretraining. We do not perform any data augmentation strategies. The base learning rate is 1.5e-4. We use AdamW \cite{loshchilov2018decoupled} to optimize the model parameters with a weight decay of 0.05. The cosine decay \cite{loshchilov2016sgdr} is applied to schedule the learning rate changes.  We only apply the RandomResizedCrop augmentation strategy. The warm-up epoch is 40 for pretraining 1,600 epochs, 20 for pretraining 800 epochs, and 10 for pretraining 400 epochs.

\textbf{Finetuning.}
We use the pre-trained visual encoder and add another classification head during the finetuning stage. The base learning rate is 1e-3. We apply the adamW \cite{loshchilov2018decoupled} optimizer and cosine decay \cite{loshchilov2016sgdr} learning rate scheduler in our implementation. We finetune the models for 100 epochs, the same as MAE. The batch size is 1024. The warm-up epoch is 5. The mixup \cite{zhang2018mixup} rate is 0.8. We also aggregate the features from all the image patches to generate the whole image representation through average pooling.

\subsection{COCO Object Detection Experiments}  
We follow ViTDet \cite{li2022exploring} and ViTAE \cite{zhang2022vitaev2} experimental settings and replace the original MAE model with our pre-trained LoMaR. We also integrate our relative positional encoding into their model. For the image resolution of 224$\times$224, we apply our pre-trained LoMaR (4 sampled windows per image + 1,600 pretraining epochs). For the image resolution of 384$\times$384, we apply our pre-trained LoMaR (9 sampled windows per image + 1,600 pretraining epochs). The image patch size is 16$\times$16. We train 25 epochs with a batch size of 64. The input image size is 1024$\times$1024.

\section{More experimental results}
\begin{figure}[t!]
\centering
 \includegraphics[width=\linewidth]
         {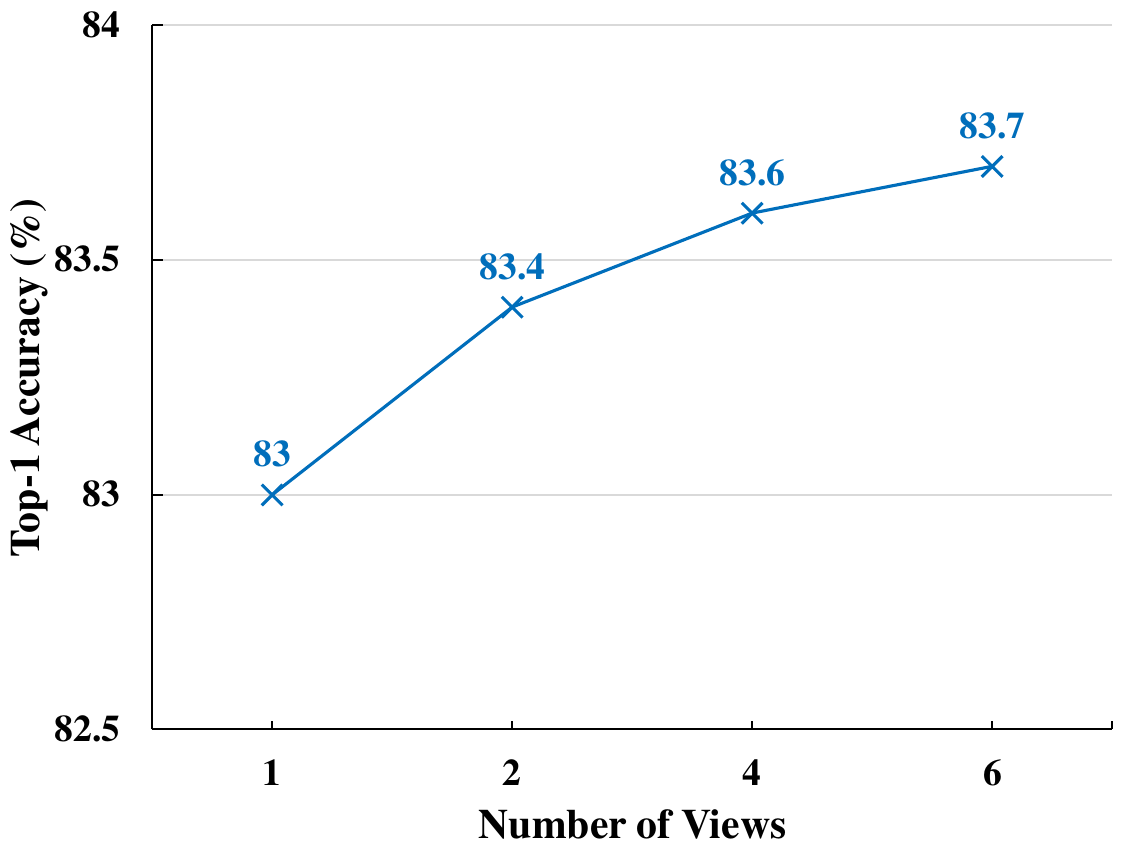}
\caption{LoMaR performance over different number of views}
\label{number_view}
\end{figure}

\subsection{Ablations on Different Number of Views.} To explore the effect of different numbers of views on the LoMaR pretraining, we additionally sample 1, 2, and 6 views per image. We follow our previous experimental setting to pretrain the model for 400 epochs and finetune the model on the ImageNet-1K dataset. The results can be found in Fig \ref{number_view}.

\noindent \textbf{RPE vs. APE.} Relative positional encoding (RPE) has been widely used in previous works, including  BEiT~\cite{beit}. 
In our LoMaR, RPE can enable our local masked reconstruction with the translation-invariant property, meaning that features of the same object under different regions would not be influenced by the different absolute positional encoding. This is especially more important for our local masked reconstruction compared to the global one. 
We also employ the RPE~\cite{wu2021rethinking} in LoMaR. We observe that it can bring 0.4 top-1 accuracy gain from 83.1 to 83.5. 
Therefore, we set RPE as our default setting for LoMaR.

\begin{table}[t!]
\begin{center}
\begin{tabular}{lccc}
\toprule
Method &  Absolute Position& Relative Position\\
\midrule
MAE & 82.5 &  83.1\\
\rowcolor{Gray2}
LoMaR  &  \textbf{83.1}& \textbf{83.5} \\
\bottomrule
\end{tabular}
\end{center}
\caption{The results of adding relative positional encoding under local masked reconstruction}
\label{ape_rpe}
\end{table}


\section{More reconstruction results}
\begin{figure*}[h]
\centering
\includegraphics[width=1.0\linewidth,height=1.2\linewidth]
                  {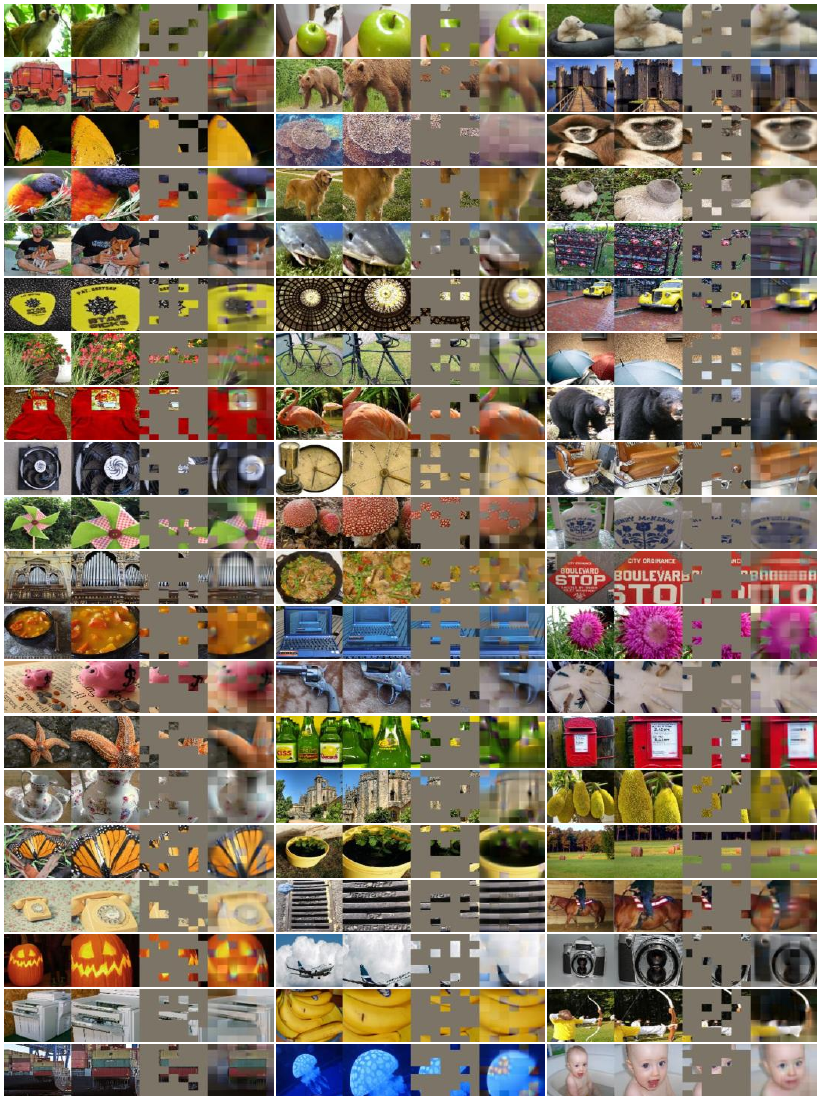}
\caption{More reconstruction examples of our pretrained model on ImageNet validation images. The masking ratio is 80\%. For each
image reconstruction figure, we split them into 4 parts: 1) the left-most is the original image. 2) the
second-left is the sampled window (7×7 patches). 3) The second-right is the masked image. 4) The
right-most is our reconstructed image}
\label{ImageNet-val}
\end{figure*}

\begin{figure*}[h]
\centering
\includegraphics[width=1.0\linewidth,height=1.2\linewidth]
                  {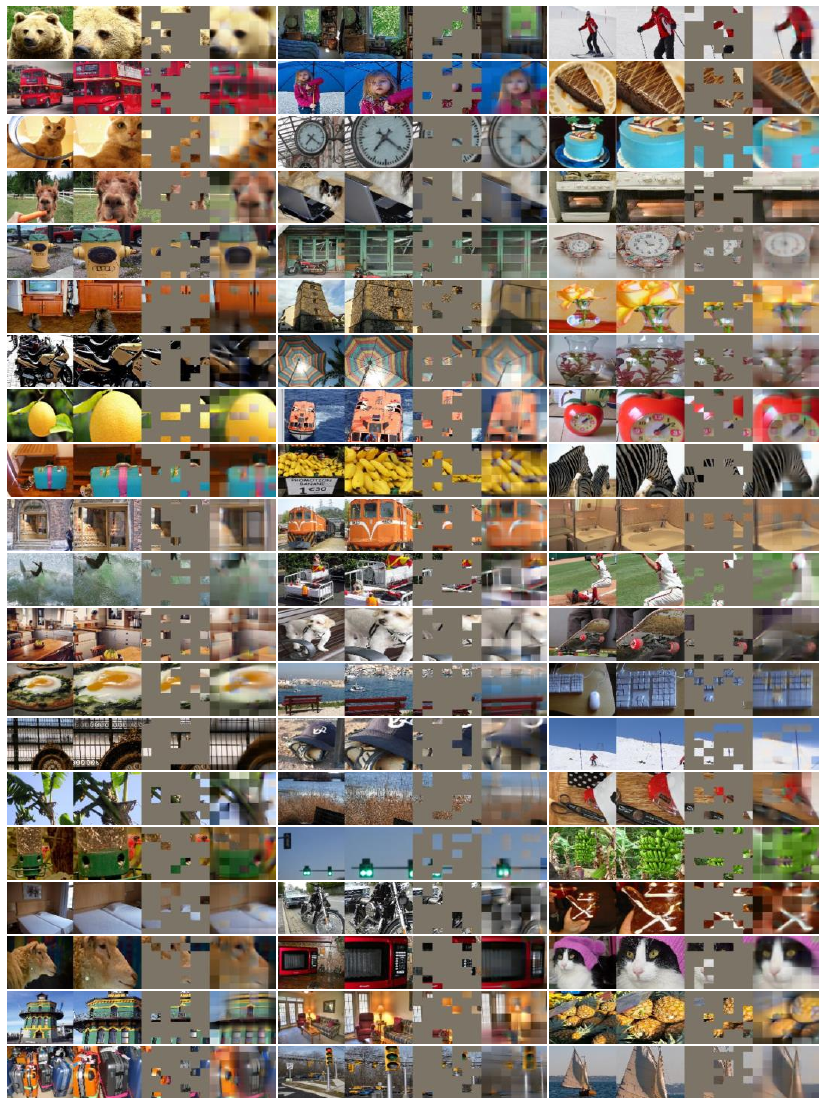}
\caption{More reconstruction examples of our pretrained model on COCO validation images. The masking ratio is 80\%. For each image reconstruction figure, we split them into 4 parts: 1) the left-most is the original image. 2) the
second-left is the sampled window (7×7 patches). 3) The second-right is the masked image. 4) The
right-most is our reconstructed image}
\label{COCO-val}
\end{figure*}

We sample more images from ImageNet \cite{imagenet} and MS COCO \cite{mscoco} and perform our local masked reconstruction. The visualization results are shown in the Fig. \ref{ImageNet-val} and \ref{COCO-val}.

\end{document}